\documentclass[10pt,twocolumn,letterpaper]{article}

\usepackage{pifont}
\usepackage{multirow}
\usepackage[table]{xcolor}
\usepackage{array}
\usepackage[most]{tcolorbox}
\usepackage{lipsum}
\usepackage[pagenumbers]{cvpr} % To force page numbers, e.g. for an arXiv version

\definecolor{cvprblue}{rgb}{0.21,0.49,0.74}
\usepackage[pagebackref,breaklinks,colorlinks,allcolors=cvprblue]{hyperref}

\def\paperID{15383} % *** Enter the Paper ID here
\def\confName{CVPR}
\def\confYear{2026}

\title{
IE-Critic-R1: Advancing the Explanatory Measurement of Text-Driven Image Editing for Human Perception Alignment}

\author{%
  Bowen Qu\textsuperscript{1} \footnotemark[1] \quad Shangkun Sun\textsuperscript{1,2} \footnotemark[1] \quad Xiaoyu Liang\textsuperscript{1} \quad Wei Gao \textsuperscript{1,2} \\
\textsuperscript{1} School of Electronic and Computer Engineering, Peking University \\
\textsuperscript{2} Peng Cheng Laboratory, China \\
\texttt{\small{\{bowenqu, sunshk, 2000017789\}@stu.pku.edu.cn, gaowei262@pku.edu.cn}} \\
\normalsize{\url{https://github.com/Coobiw/IE-Critic-R1}}
}

\begin{document}

\maketitle

\footnotetext[1]{\quad$^*$ Equal Contribution}

\begin{abstract}
Recent advances in text-driven image editing have been significant, yet the task of accurately evaluating these edited images continues to pose a considerable challenge. Different from the assessment of text-driven image generation, text-driven image editing is characterized by simultaneously conditioning on both text and a source image. The edited images often retain an intrinsic connection to the original image, which dynamically change with the semantics of the text. However, previous methods tend to solely focus on text-image alignment or have not well aligned with human perception. In this work, we introduce the Text-driven Image Editing Benchmark suite (IE-Bench) to enhance the assessment of text-driven edited images. IE-Bench includes a database contains diverse source images, various editing prompts and the corresponding edited results from different editing methods, and nearly 4,000 samples with corresponding Mean Opinion Scores (MOS) provided by 15 human subjects. Furthermore, we introduce IE-Critic-R1, which, benefiting from Reinforcement Learning from Verifiable Rewards (RLVR), provides more comprehensive and explainable quality assessment for text-driven image editing that aligns with human perception. Extensive experiments demonstrate IE-Critic-R1's superior subjective-alignments on the text-driven image editing task compared with previous metrics. Related data and codes are available to the public.
\end{abstract}    
\section{Introduction}
\label{sec:intro}

\begin{figure*}[t]
\centering
\includegraphics[width=2.1\columnwidth]{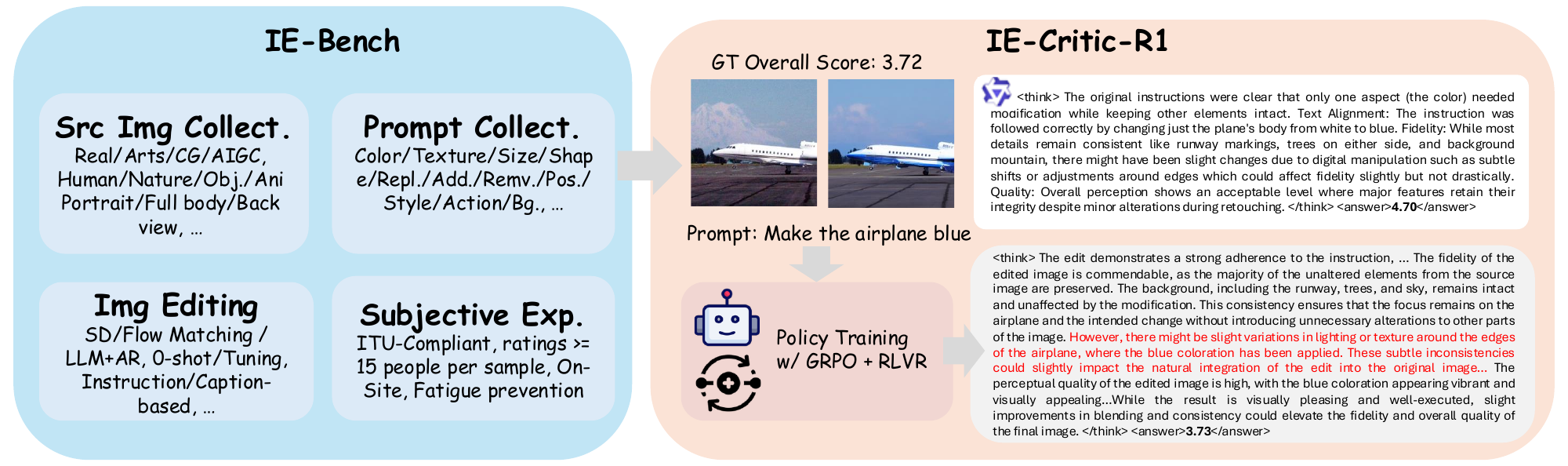} % Reduce the figure size so that it is slightly narrower than the column. Don't use precise values for figure width.This setup will avoid overfull boxes.
\caption{Overview of IE-Bench and IE-Critic-R1. Compared to the Qwen Baseline, the IE-Critic-R1 trained through RLVR design has stronger thinking ability.}
\label{fig:1}
\end{figure*}

Text-driven image editing~\cite{instructpix2pix, masactrl, magicbrush, sine, dac} has attracted significant attention in recent years. However, there is currently no well-established metric for evaluating the results of image editing. Objective metrics such as CLIP score~\cite{clip}, DINO score~\cite{dino}, LPIPS score~\cite{lpips}, and SSIM~\cite{ssim} tend to evaluate image quality from a single perspective, such as text-image consistency or the correlation between the source and edited images. These metrics, however, do not provide an overall evaluation, nor do they align well with human perception. Previous studies~\cite{t2iqa, hps, pickscore, closing} have shown that these metrics can significantly differ from human judgment in practical applications.

In recent years, some metrics aligned with human perception, such as HPS scores~\cite{hps, hpsv2, hpsv3}, Pick score~\cite{pickscore}, and ImageReward~\cite{imagereward}, have made effective progress in evaluating text-to-image generation tasks by collecting human visual feedback. However, these methods focus only on individual images and text, which differs from the setting of image editing tasks. Unlike text-driven image generation, text-driven image editing also takes a source image as input. The edited result is expected to differ from the source image, but there is also a certain degree of correspondence. Modeling this relationship is crucial for evaluation: in some cases, the edited result is expected to retain semantic information related to the original image. If only the edited image output is considered, this aspect would be missed, which is an issue to be well-addressed.

However, modeling this relationship is a challenging task. The connection between source and target images changes dynamically depending on the text context. For example, a stylistic instruction like "make it a claymation style" may drastically alter the structure, texture, and lines of the original image, whereas a replacement instruction like "replace the cat with a dog" will directly alter the semantic content, and thus a large difference between the source and target images is expected. On the other hand, an instruction like "remove her earrings" is expected to retain most of the identity information of the original character. Therefore, a multi-modal method that can dynamically model the source-target relationship is urgently needed.

In this work, we propose the Text-driven Image Editing Benchmark (IE-Bench) to improve the alignment between evaluation metrics for text-driven image editing and human perception. We first introduce IE-Bench, a database containing various source-prompt-target cases and their corresponding Mean Opinion Scores (MOS). We collect diverse real-world, CG, AIGC, and art painting images from different sources. Following previous works~\cite{huang}, we manually design diverse editing instructions for each image, covering aspects such as structural changes (e.g., shape, size), style changes (e.g., texture, color), and semantic changes (e.g., pose, action, addition, replacement, deletion). We then apply multiple methods to generate diverse edited results. Finally, we assemble 15 human participants from various backgrounds to provide subjective ratings. These ratings follow the ITU standard~\cite{itu}, and the detailed process is described in Section~\ref{sec:tie_data}.

To address these challenges, we introduce IE-Critic-R1, a comprehensive and explainable quality assessment method for text-driven image editing. We firstly leverage the multi-dimensional human annotated scores in IE-Bench to prompt GPT-4o~\cite{gpt-4o} for the Chain-of-Thought (CoT) reasoning data, which is then mixed with direct scoring supervised fine-tuning (SFT) data to gain IE-Critic-CoT. Building upon this strong cold-start model, we apply reinforcement learning with carefully designed reward function, successfully observing the "R1 Moment" where response length increases with training steps (as shown in Figure~\ref{fig:response-length-curve}), ultimately yielding IE-Critic-R1. Our main contributions are as follows: (1) \textbf{IE-Bench}: We construct a large-scale benchmark dataset containing diverse source-prompt-target triplets accompanied by multi-dimensional human annotations (text alignment, fidelity, quality, and overall scores). (2) \textbf{IE-Critic-CoT}: We propose a CoT data synthesis strategy that prompts GPT-4o with multi-dimensional scores from IE-Bench to generate reasoning processes. Through mixing CoT data with direct scoring data during SFT phase, the model learns both comprehensive reasoning and direct prediction capabilities. (3) \textbf{IE-Critic-R1}: We introduce a Reinforcement Learning from Verifiable Reward (RLVR) framework with optimal reward design. Our approach successfully identifies the "R1 Moment" in image editing quality assessment, where models generate increasingly longer and more detailed reasoning as training progresses. (4) \textbf{Superior Performance}: The comprehensive and explainable CoT reasoning enables IE-Critic-R1 to achieve state-of-the-art performance on IE-Bench and AGIQA-3k~\cite{agiqa}, demonstrating substantial improvements in alignment with human perception.

% Building on IE-Bench, we further develop IE-Critic-R1, a source-aware multi-modal assessment method for text-driven image editing. IE-Critic-R1 introduces an effective way to model the source-target relationship in image editing tasks and provides a comprehensive and explainable evaluation by considering multiple dimensions, such as the visual quality, visual-text alignment, and the connection between the source and target images. Extensive experiments demonstrate its superior alignment with human perception compared to traditional methods. Our contributions are as follows: (1) We introduce IE-Bench, which collects diverse source-prompt-target cases from various sources along with corresponding MOS reflecting human subjectives. (2) We propose IE-Critic-CoT, which further develops IE-Critic-R1, which incorporates both the source-target relationship and multi-modal text input, offering an effective way to dynamically model the source-target relationship in response to multi-modal inputs. (3) IE-Critic-R1 demonstrates a significant advantage in subjective alignment compared to previous IQA methods, effectively showcasing the potential of modeling the source-target relationship during evaluation.

\section{Related Work}
\label{sec:related_work}

\subsection{Measurements for Image Editing}
Currently, the evaluation methods commonly used in text-driven image editing include several objective metrics~\cite{clip,lpips,ssim,fid, vgeneval}, as well as some quality assessment methods~\cite{hps,hpsv2,imagereward,pickscore, vebench, ntire, qu2024exploring, crave} aligned with human feedback. CLIP-V~\cite{clip} calculates the cosine similarity between each edited image and the source image, while CLIP-T measures the relationship between the result and the given text prompt. MSE and SSIM~\cite{ssim} represent the variance in pixels and overall structure between the edited image and the source image. FID~\cite{fid} calculates the Fréchet Distance between two images. The DINO score~\cite{dino} measures semantic consistency and calculates feature variance. However, these individual metrics often assess the editing results only from a single dimension. For instance, they either measure only the source-target relationship or the visual-text connection, without aligning with human subjective perception. PickScore~\cite{pickscore} estimates alignment with human preferences via a CLIP-style model fine-tuned on human preference data. ImageReward~\cite{imagereward}, and HPS scores~\cite{hps, hpsv2, hpsv3}, evaluate natural images from aesthetic and technical distortion perspectives. Despite effective scoring based on human feedback training, these methods do not consider the inherent relationship between edited results and the source image, and some traditional IQA methods like~\cite{dbcnn} do not model the alignment between text and image. Currently, there is still a lack of an appropriate metric to evaluate edited results based on the source image and the editing prompts.

\subsection{Image Editing Methods and Datasets}

With the development of deep learning methods~\cite{skflow, opendmc, streamflow,adaptive,closing}, pre-trained text-to-image diffusion models~\cite{dreambooth, sd, ddpm} have demonstrated strong capabilities in image editing tasks. Recent instruction-based methods~\cite{instructpix2pix, magicbrush, pnp, masactrl} enable users to edit images through natural language instructions by modifying attention mechanisms or leveraging large-scale vision-language datasets. Advanced approaches~\cite{bagel, sine, dac} further enhance editing performance by incorporating large language models or test-time fine-tuning. Despite these advances, evaluating editing quality remains challenging. Traditional evaluation relies on human annotators conducting subjective preference experiments~\cite{smartedit, refcoco-edit}, but such results are difficult to reproduce due to inconsistent experimental setups. While recent efforts~\cite{instructpix2pix, sine} have introduced standardized benchmark datasets, these datasets typically lack subjective feedback scores (e.g., MOS) and have limited scenario coverage, requiring researchers to conduct new subjective experiments or rely on objective metrics that may not fully align with human perception.

\subsection{Reinforcement Learning for VLMs Reasoning}
With the recent success of DeepSeek-R1~\cite{deepseek-r1}, many works focus on exploring the reasoning ability of Large Lanuge Models (LLM)~\cite{dapo, raft, skywork-or1, kimik2} and Multimodal Large Langeuage Models (MLLM)~\cite{mmeureka, mmr1, vlmr1,mammothvl, aria, flow4agent}. Researchers use Reinforcement Learning from Verifiable Reward (RLVR) methods to improve many visual reasoning tasks, such as math reasoning~\cite{mmeureka, kimi-vl}, chart reasoning ~\cite{mmr1, chartmoe} and visual grounding tasks~\cite{vlmr1}. The characteristic "R1 moment" is reflected in a steady increase in both accuracy and response length as the number of training steps increases. This phenomenon happens on math reasoning tasks usually. As for visual quality assessment, Q-Insight and VQ-Insight focus on image and video domain respectively. Without a specially prepared cold-start stage, it is difficult for them to find the growth of response length.

\begin{figure*}[t]
\centering
\includegraphics[width=1.0\linewidth]{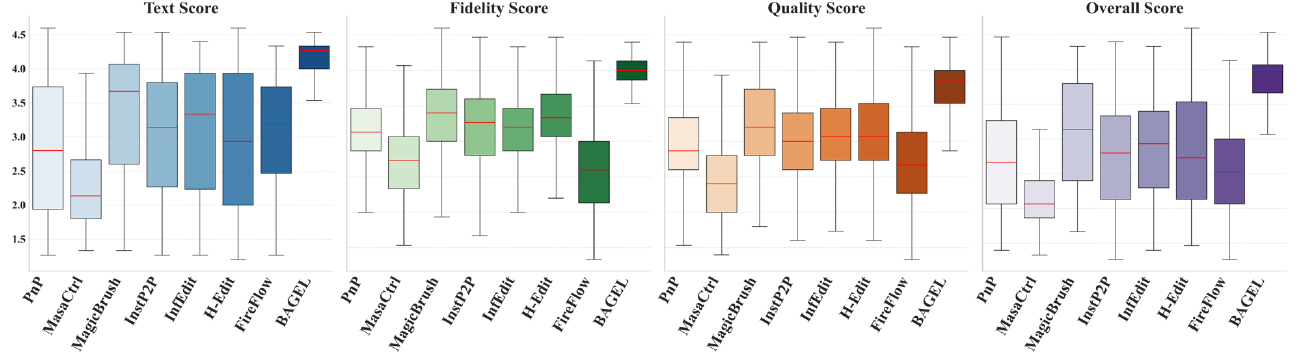} % Reduce the figure size so that it is slightly narrower than the column. Don't use precise values for figure width.This setup will avoid overfull boxes.
\caption{Score distributions of various editing methods. The central line represents the median, and the whiskers extend to the minimum and maximum values.}

\label{fig:mos}
\end{figure*}

\section{Text-driven Image Editing Database}
\label{sec:tie_data}
The collection of IE-Bench involves four primary stages: source image collection, prompt selection and execution of image editing methods, and subjective experiments. % We will elaborate on each part in subsequent sections with corresponding analyses.

% \begin{figure*}[t]
% \centering
% \includegraphics[width=2.0\columnwidth]{figs-tie/data.pdf} % Reduce the figure size so that it is slightly narrower than the column. Don't use precise values for figure width.This setup will avoid overfull boxes.
% \caption{Collection of source images. (a) Sources of images. (b) Categories of images. (c) Content of images. (d)~(f) denotes fine-grained classification of human images, including classification of camera, pose, and actions, respectively.} %【视频数据的来源(哪些数据集)的饼状图、真实/图形学/AIGC的饼状图、Motion类别构成图、以及(人、动物、风景、物体等)内容类别构成饼状图】. }
% \label{fig:src_image_collection}
% \end{figure*}

\subsection{Source Image Collection}
IE-Bench compiles a rich and varied collection of source images to facilitate a more robust evaluation of image editing quality. In addition to real-world scenes, the dataset incorporates computer-generated imagery, text-driven generated images, and artistic creations. Real-world scenes, given their widespread application, constitute the largest portion of the dataset. Unlike previous efforts, IE-Bench refrains from random sampling of images from large datasets to avoid complications related to copyright, watermarks, and image resolution. Instead, to cover as many different content subjects, action categories, and scenarios as possible, IE-Bench manually selected diverse images from four datasets: ADE~\cite{ade}, WIKIArt~\cite{wikiart}, COCO~\cite{coco}, ReasonEdit~\cite{smartedit}, and other Internet sources. Instead of sampling from large-scale dataset, we first confirmed the distribution of data sources (CG, AIGC, real-world datasets, and artist-created works), then iterated through each sample to tag attributes such as ``[Landscape/Object/Animal/Human]" and ``[action type]". Samples of the same type (e.g., Landscape / Object / Animal or Human action) or with insufficient resolution were skipped until the dataset was fully reviewed or the relevant categories were sufficiently populated. Smaller datasets were prioritized for review, and additional samples were drawn from larger datasets. Finally, suitable images were selected from the internet for supplementation. For instance, while many datasets included landscapes like grasslands and snowy mountains, scenes such as auroras, lava flows, and lightning were less represented. Similarly, although current datasets provide a rich variety of action categories, there remains a limited number of actions with significantly varied patterns. Ultimately, we collected over 100 human action types, as illustrated in the \emph{supplements}. %Figure~\ref{fig:src_image_collection}. 
For simplicity, the remaining categories were grouped under "others," with detailed explanations provided in the supplementary materials. Ultimately, a diverse set of \emph{301} source images was gathered, encompassing a wide range of content. The origins, contents, and category distributions of these images are illustrated in the \emph{supplements}.
% Figure~\ref{fig:src_image_collection}. 
Each image was resized to ensure the shorter side was 512 pixels, preserving the original aspect ratios.

\subsection{Prompt Selection}
\begin{figure*}[htbp]
\centering
\includegraphics[width=1.5\columnwidth]{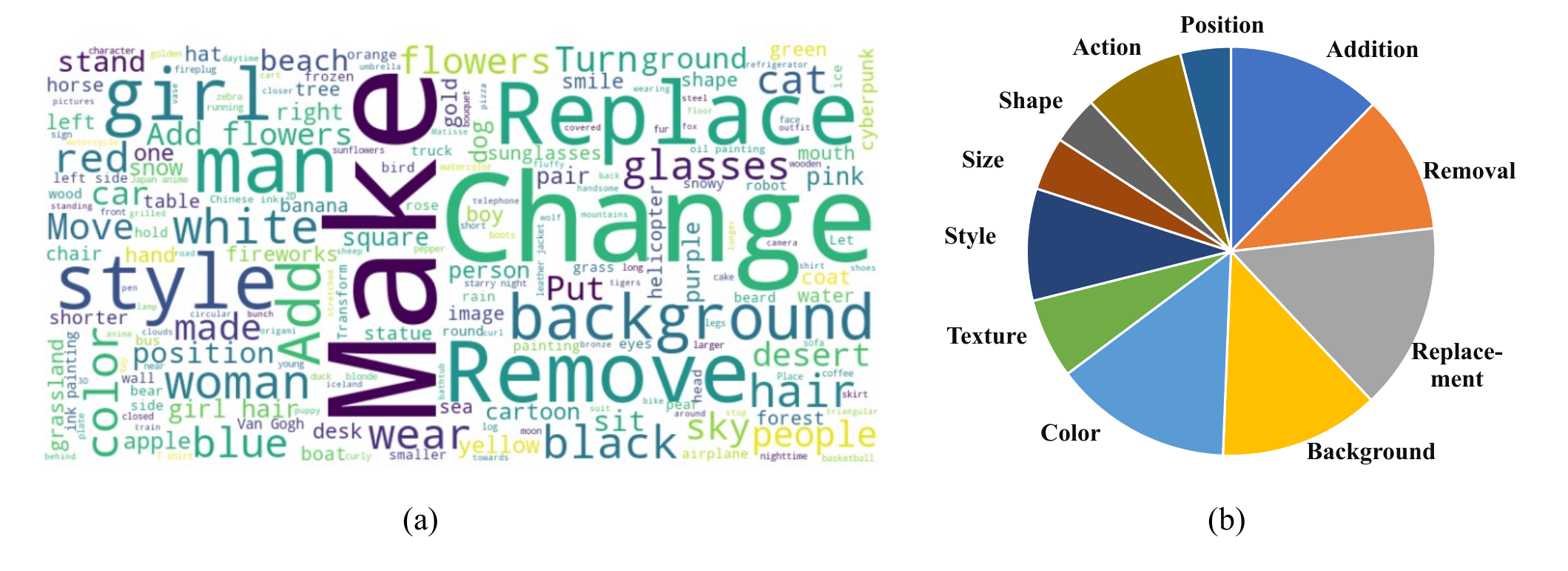} % Reduce the figure size so that it is slightly narrower than the column. Don't use precise values for figure width.This setup will avoid overfull boxes.
\caption{Statistics of IE-Bench prompts. (a) Word cloud of IE-Bench DB prompts. (b) Proportion of different types}

\label{fig:prompt_info}
\end{figure*}

Building upon prior works~\cite{prompt}, we categorize image editing prompts into three primary types: (1) Style editing, involving adjustments to color, texture, or overall ambiance. (2) Semantic editing, which encompasses background modifications and localized edits such as adding, replacing, or removing specific objects. (3) Structural editing, including changes to object size, pose, or motion. To guarantee the variety and specificity of the prompts, we created tailored prompts for each image, as shown in Figure~\ref{fig:prompt_info}.

\subsection{Image Editing}
We selected 8 diverse image editing techniques. To ensure a balanced quality distribution among the edited images, our selection includes both state-of-the-art models and earlier approaches. We also incorporated methods based on various foundational models, ranging from SD 1.4 to SD 2.1~\cite{sd}, as well as DiT-based methods like FLUX~\cite{flux}, and unified modeling approaches based on LLMs~\cite{qwen2.5}, to enhance the diversity of editing outcomes. These methods differ not only in their backbone architectures but also in the editing techniques they employ. For example, Bagel~\cite{bagel} is a generation-understanding unified model based on LLM capabilities, FireFlow~\cite{fireflow} relies on inversion, Instruct-Pix2Pix~\cite{instructpix2pix} uses implicit learning, Prompt-to-Prompt~\cite{pnp} and MasaCtrl~\cite{masactrl} are based on attention control, and H-Edit~\cite{h-edit} utilizes reverse-time bridge modeling. This variety of methods ensures diverse generation results and provides a comprehensive reflection of the current state of image editing models. Furthermore, to diversify the edited content, we included both zero-shot methods and those that require fine-tuning. The specifics of these methods are outlined in the \emph{supplements}. % Table~\ref{tab:editing_methods}.

% \begin{table*}[htb]
% \centering
% \scalebox{1.0}{
% \begin{tabular}{llccc}
% \toprule
% Model & Time & 0-shot & Type & Backbone. \\ \midrule
% Instruct-Pix2Pix~\cite{instructpix2pix}      &  CVPR'23   &    \ding{55}   &   Instruction-based  &    SD v1-4       \\ \midrule
% Prompt-to-Prompt~\cite{pnp}      &  CVPR'23    &    \ding{51}       &     Description-based       &   SD v1-5         \\ \midrule
% MagicBrush~\cite{magicbrush}      &  NeurIPS'23    &    \ding{55}       &     Instruction-based       &   SD v1-5    \\ \midrule
% MasaCtrl~\cite{masactrl}      & ICCV'23     &   \ding{51}          &    Description-based             &  SD v1-4          \\ \midrule
% InfEdit~\cite{infedit}      & CVPR'24     &  \ding{51}           &  Description-based      &    SD v2-1       \\ \midrule
% H-Edit~\cite{h-edit} & CVPR' 25 & \ding{51} & Description-based & SD v1-4 \\ \midrule
% FireFlow~\cite{fireflow} & ICML' 25 & \ding{51} & Description-based & FLUX-1-dev \\ \midrule
% BAGEL~\cite{bagel} & Tech Report' 25 & \ding{51} & Unified & Qwen2.5-LLM \\
% \bottomrule
% \end{tabular}
% }
% \caption{Collection of the editing models.}
% \label{tab:editing_methods}
% \end{table*}

\subsection{Subjective Study}
In accordance with ITU standards~\cite{itu}, subjective experiments necessitate a minimum of 15 participants to ensure result variance remains within acceptable limits. 
We strictly adhere to this requirement, ensuring each case is evaluated by 15 participants with diverse professional backgrounds. 
% These individuals were tasked with evaluating edited images based on text-image consistency, source-target fidelity, and overall quality, relying on their subjective judgments. 
% All participants were over 18 years old, held at least an undergraduate degree, and had varied professional experiences, including business, engineering, science, and law, ensuring their ability to make independent judgments. 
Prior to the experiment, participants underwent in-person training, where they were shown examples of high-quality and poor edits not included in the dataset. During the assessment, each participant evaluated all image samples, with a mandatory 5-minute break every 15 minutes to minimize fatigue. 
Our procedures were consistent with those in previous subjective studies, such as~\cite{t2vqa, agiqa, keimel2012tum}. Text-image consistency was defined as the extent to which the edited content aligns with the provided prompt. Source-target fidelity measures how well the edited image retains a connection to the original. Participants rated these aspects on a 1 to 10 scale during their evaluations. Consistent with prior research~\cite{e-bench,t2vqa,liqe}, we employed Z-score normalization for the raw MOS values. Following the collection of raw Mean Opinion Scores (MOS), we applied Z-score normalization to account for inter-subject variability, using the formula:

\begin{align}
    Z_{m,i} = \frac{X_{m,i} - \mu(X_i)}{\sigma(X_i)},
\end{align}

where $X_{m,i}$ denotes the raw MOS and $Z_{m,i}$ the Z-score for the $m$-th image evaluated by the $i$-th participant. Here, $\mu(\cdot)$ and $\sigma{(\cdot)}$ represent the mean and standard deviation, respectively, and $X_{i}$ is the set of all MOS scores from participant $i$. We also filter outliers following BT.500~\cite{bt500}.

\subsection{Statistcs Analysis}
We performed a comprehensive analysis of the IE-Bench dataset and examined the evaluation scores for the image editing outcomes produced by each model on the gathered images, as depicted in Fig.~\ref{fig:mos}. Overall, we found that the median scores of different models mainly fluctuate within the range of [2, 4.5] in the 1-5 scoring interval, with most models scoring around 3. Specifically, models generally have higher scores in the Text Alignment dimension, with the median scores in this dimension ranging between [2, 4.5]. The scores for Quality and Fidelity tend to be lower, with median scores concentrated in [2,4]. The highest and lowest scores across different dimensions do not differ significantly. Nevertheless, for the same model, there may be obvious differences in scores across different dimensions. For example, MagicBrush excels in Text Score, but its advantages in Fidelity Score and Quality Score are relatively modest. Impressively, we found that BAGEL has outstanding scores across all dimensions with relatively small variance, indicating the great potential of image editing methods based on unified modeling. We also found that it exhibits significant advantages in both Text and Fidelity Score performance. However, in terms of quality, while it still leads compared to other diffusion-based or flow matching-based methods, the advantage is less pronounced, which may indicate potential improvements for unified models.

\section{Method}

To provide a comprehensive and explainable quaility assessment of edited images, we introduce our proposed multi-dimensional Chain-of-Thoughts (CoT) data synthesis and two-stage training strategy. We first use the multi-dimensional human annotated scores from IE-Bench to generate the CoT reasoning process by prompting GPT-4o~\cite{gpt-4o}. We employ the Supervised Fine-Tuning (SFT) on CoT data as a cold start phase for the subsequent RL stage.

\subsection{IE-Critic-CoT}

To enable the model to learn comprehensive reasoning capabilities for image editing quality assessment, we leverage GPT-4o to generate CoT reasoning data based on the multi-dimensional human annotations in IE-Bench. Specifically, we provide GPT-4o with the source image, edited image, editing instruction, and the four-dimensional human annotated scores (text alignment, fidelity, quality, and overall). GPT-4o is prompted to expand the human comments by comprehensively analyzing the given images and editing instruction, according to the given clues.

This generated CoT data serves as the cold-start for Qwen-2.5-VL-7B-Instruct~\cite{qwen-2.5-vl}. Through SFT on this data, the model learns to compare the source and edited images, understand the editing instruction, assess the consistency between the instruction and the edit, and evaluate the visual quality of the edited image. The reasoning process guides the model to systematically analyze each dimension before arriving at a comprehensive overall score, thereby establishing a foundation for explainable quality assessment.

Furthermore, we decompose the image editing quality assessment task into two sub-tasks: reasoning then scoring. During SFT phase, we mix data from both the full task (CoT reasoning followed by score prediction) and the scoring-only task (direct score prediction without explicit reasoning). This task decomposition and data recipe plays a crucial role in the subsequent RL stage. By learning both reasoning-then-scoring and direct scoring capabilities, the model develops the ability to generate stronger positive trajectories, which raises the upper bound of reinforcement learning and enables more effective alignment with human assessment.

\subsection{IE-Critic-R1}
After the SFT stage, the model demonstrates an enhanced capability to generate reasoning processes from multiple aspects and provide relatively precise overall scores. However, the stability of the reasoning process and the final performance on out-of-domain tasks remain limited. Although CoT SFT teaches the model to imitate the behavior of GPT-4o, the model also needs to learn from its own outputs in real-world environments through reinforcement learning. Therefore, we introduce IE-Critic-R1, a Reinforcement Learning from Verifiable Reward (RLVR) stage that leverages Group Relative Policy Optimization (GRPO)~\cite{deepseek-r1} for futher improvement on generalization.

\subsubsection{Group Relative Policy Optimization} 
GRPO is a policy gradient method that optimizes the model by comparing its multiple output trajectories generated from the same prompt. GRPO foregoes the value model, instead, using group scores to estimate the baseline, which is friendly to training resources. For each input prompt $x$, we sample $G$ output trajectories $\{y_1, y_2, \ldots, y_G\}$ from the current policy $\pi_\theta$. Each trajectory $y_i$ is evaluated by the reward function $r_i$ that measures the distance between the predicted score and the ground-truth human annotation. The group-based advantage for trajectory $y_i$ is computed as:
\begin{align}
    A_i = \frac{r_i - \bar{r}}{\sigma_r},
\end{align}
where $\bar{r} = \frac{1}{G}\sum_{j=1}^{G} r_j$ is the mean reward of this group and $\sigma_r$ is the standard deviation of the group rewards. The GRPO objective with importance sampling and clipping~\cite{ppo} is formulated as:
\begin{align}
    \mathcal{J}_{\text{GRPO}} &= \mathbb{E}_{x, \{y_i\}_{i=1}^{G}} \left[ \frac{1}{G}\sum_{i=1}^{G} \min\Big(ratio_i, \nonumber \right. \\
    &\quad \left. \text{clip}(ratio_i, 1-\epsilon, 1+\epsilon)\Big) A_i \right] \nonumber \\
    &\quad - \beta \cdot \mathbb{E}_x[\text{KL}(\pi_\theta || \pi_{\text{sft}})],
\end{align}
where $ratio_i = \frac{\pi_\theta(y_i|x)}{\pi_{\text{old}}(y_i|x)}$ is the importance sampling ratio between the current policy and the old policy, $\epsilon$ is the clipping threshold, $\pi_{\text{sft}}$ is the reference policy (the SFT model), and $\beta$ controls the strength of the KL regularization to prevent the policy from deviating too far from the reference model. In order for better performance, we set $\beta$ to 0.

\subsubsection{Reward Design}

The reward function plays a crucial role in guiding the policy optimization process. We design a composite reward function that considers both the format correctness and the accuracy of the predicted score. The overall reward is formulated as:
\begin{align}
    r = r_{\text{acc}} + \lambda \cdot r_{\text{fmt}},
\end{align}
where $r_{\text{acc}}$ measures the accuracy of the predicted score against the human-annotated overall score, $r_{\text{fmt}}$ ensures proper output formatting, and $\lambda$ is a coefficient to control the contribution of the format reward. Here, we set $\lambda$ to 1.

\textbf{Format Reward.} The format reward $r_{\text{fmt}}$ ensures that the model produces outputs in the expected structure. Specifically, we require the model to enclose its final score prediction within designated markers. The format reward is binary: $r_{\text{fmt}} = 1$ if the output follows the correct format, and $r_{\text{fmt}} = 0$ otherwise. This encourages the model to maintain consistent and parsable output structure throughout training.

\textbf{Accuracy Reward.} The accuracy reward $r_{\text{acc}}$ measures how close the predicted score $s_{\text{pred}}$ is to the ground-truth score $s_{\text{gt}}$. The $s_{\text{pred}}$ and $s_{\text{gt}}$ range from 1 to 5. We explore four types of shaped reward functions that smoothly decay as the prediction error increases. Table~\ref{tab:reward_functions} summarizes the mathematical forms of these reward functions, where $d = |s_{\text{pred}} - s_{\text{gt}}| / 4$ denotes the normalized absolute error (scores range from 0 to 4).

\begin{table}[h]
\centering
\caption{Four types of accuracy reward functions.}
\label{tab:reward_functions}
\begin{tabular}{l|c}
\toprule
Function Type & Mathematical Form \\
\midrule
$\ell_1$ (Linear) & $r_{\text{acc}} = 1 - \alpha \cdot |s_{\text{pred}} - s_{\text{gt}}|$ \\
$\ell_2$ (Quadratic) & $r_{\text{acc}} = 1 - \alpha \cdot (s_{\text{pred}} - s_{\text{gt}})^2$ \\
Laplacian & $r_{\text{acc}} = \exp\left(-\frac{d}{\tau}\right)$ \\
Gaussian & $r_{\text{acc}} = \exp\left(-\frac{d^2}{2\sigma^2}\right)$ \\
\bottomrule
\end{tabular}
\end{table}

\begin{figure}[t]
\centering
\includegraphics[width=1.\columnwidth]{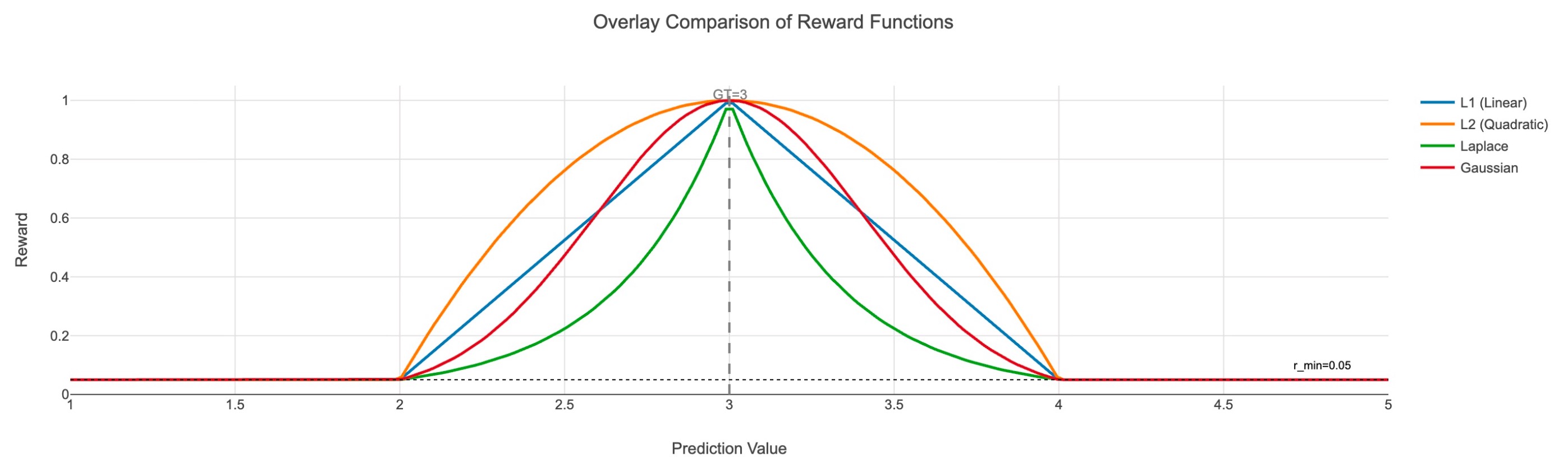} % Reduce the figure size so that it is slightly narrower than the column. Don't use precise values for figure width.This setup will avoid overfull boxes.
\caption{Overlay Comparison of Reward Functions.}
\label{fig:reward_functions}
\end{figure}

For all four reward functions, the shape parameters ($\sigma$, $\tau$, or $\alpha$) are determined by two hyperparameters: $r_{\text{min}}$ (the reward floor value) and $d_0$ (the error threshold at which the reward decays to $r_{\text{min}}$). Specifically, for Gaussian, $\sigma = d_0 / \sqrt{2\ln(1/r_{\text{min}})}$; for Laplacian, $\tau = d_0 / \ln(1/r_{\text{min}})$; for $\ell_1$ and $\ell_2$, $\alpha = (1 - r_{\text{min}}) / d_0$ and $\alpha = (1 - r_{\text{min}}) / d_0^2$ respectively. To prevent reward sparsity, we apply a floor operation $r_{\text{acc}} = \max(r_{\text{acc}}, r_{\text{min}})$ to all functions. Here, we set $d_0$ and $r_{min}$ to 1 and 0.05, respectively. Through following empirical evaluation, we find that the $\ell_1$ reward function achieves the best on both encouraging longer reasoning contents and achieving strong performance on image editing quality assessment. Figure~\ref{fig:reward_functions} illustrates the decay patterns of these four reward functions.

\section{Experiments}

\subsection{Implementation Details}
We split IE-Bench into training and validation sets with an 4:1 ratio, and the overall MOS scores are normalized to the range [1,5]. We use Qwen-2.5-VL-7B-Instruct as the intial model and perform SFT using LLaMA-Factory~\cite{llama-factory} to get IE-Critic-CoT. Then we use the IE-Critic-CoT model to conduct the RL stage, adopting the Easy-R1~\cite{easyr1} framework based on verl~\cite{verl} for training. The SFT stage is trained for 180 steps, and the RL stage is trained for 200 steps. For these two stages, the learning rates are set to 1e-5 and 1e-6 respectively. The training processes for SFT and RL require 1.5 and 18 hours. Both IE-Critic-CoT and IE-Critic-R1 are trainer on a system with 8 A100 (80GB) GPUs.

\subsection{Evaluation Metrics}
Consistent with prior studies~\cite{IP-IQA,t2vqa,dbcnn}, we use three evaluation metrics: Spearman’s Rank Order Correlation Coefficient (SROCC), Pearson’s Linear Correlation Coefficient (PLCC) and MainScore, which is $\frac{PLCC + SROCC}{2}$. 

\begin{table}[t]
\centering
\caption{Quantitative comparison of different methods under two evaluation settings on IE-Bench validation set.}
\label{tab:ie_bench}
\resizebox{\columnwidth}{!}{
\begin{tabular}{c|c|ccc}
\toprule
\textbf{Setting} & \textbf{Method} & \textbf{PLCC} $\uparrow$ & \textbf{SROCC} $\uparrow$ & \textbf{MainScore} $\uparrow$ \\
\midrule
\multirow{7}{*}{\shortstack[c]{\textbf{Edited}\\[3pt]\textbf{Image}\\[3pt]\textbf{Only}}}
& NIMA~\cite{nima} & 0.4588 & 0.4719 & 0.4653 \\
& ImageReward~\cite{imagereward} & 0.4768 & 0.4741 & 0.4755 \\
& CLIP-IQA~\cite{clipiqa} & 0.5159 & 0.5127 & 0.5143 \\
& Q-Align~\cite{q-align} & 0.5225 & 0.5196 & 0.5211 \\
& TOPIQ~\cite{topiq} & 0.5746 & 0.5642 & 0.5694 \\
& IP-IQA~\cite{IP-IQA} & 0.5882 & 0.5716 & 0.5799 \\
& Qwen-2.5-VL-7B~\cite{qwen-2.5-vl} & 0.6418 & 0.6270 & 0.6344 \\
\midrule
\multirow{4}{*}{\shortstack[c]{\textbf{Full}\\\textbf{Context}}}
& Q-Align~\cite{q-align} & 0.6105 & 0.5982 & 0.6044 \\
& Qwen-2.5-VL-7B~\cite{qwen-2.5-vl} & 0.8225 & 0.8191 & 0.8208 \\
& \cellcolor{blue!5}IE-Critic-CoT (Ours) & \cellcolor{blue!5}0.8384 & \cellcolor{blue!5}0.8224 & \cellcolor{blue!5}0.8304 \\
& \cellcolor{blue!5}IE-Critic-R1 (Ours) & \cellcolor{blue!5}\textbf{0.8693} & \cellcolor{blue!5}\textbf{0.8629} & \cellcolor{blue!5}\textbf{0.8661} \\
\bottomrule
\end{tabular}
}
\end{table}

\subsection{Quantitative Results}
Our results are compared against advanced evaluation metrics in image editing, including IQA methods~\cite{nima, topiq, clipiqa, IP-IQA} and MLLM-based methods~\cite{q-align, qwen-2.5-vl}. We conduct experiments under two different input settings:
\textbf{Setting 1: Edited Image Only.} In this setting, the model receives only the edited image and its corresponding caption (if the model supports) as input. 
\textbf{Setting 2: Full Context.} In this setting, the model is provided with the source image, edited image, and editing instruction. This setup enables the model to compare the source and edited images, understand the editing intent, and evaluate how well the edit aligns with the instruction. This setting better reflects the complete image editing quality assessment scenario, as shown in Table~\ref{tab:ie_bench}. 

When the model only receives the edited image, traditional IQA methods such as NIMA and CLIP-IQA achieve limited performance, with MainScore below 0.52. MLLM-based methods demonstrate stronger capability, with Qwen-2.5-VL-7B reaching a MainScore of 0.6344. This suggests that MLLMs possess better understanding of image content and quality compared to traditional IQA metrics. However, the overall performance under this setting remains relatively modest, indicating the challenge of assessing editing quality without access to the source image and editing instruction. When provided with full context (source image, edited image, and prompts), all MLLM-based methods show substantial performance improvements. Qwen-2.5-VL-7B achieves a MainScore of 0.8208, representing a significant gain of 0.1864 compared to Setting 1. % This demonstrates the importance of incorporating editing context for accurate image editing quality assessment. 
Our IE-Critic-CoT model, trained with CoT reasoning data generated by GPT-4o, further improves the MainScore to 0.8304, validating the effectiveness of our multi-dimensional reasoning approach.
IE-Critic-R1, which undergoes RL training on top of IE-Critic-CoT, achieves the best performance across all metrics with a MainScore of 0.8661. 
% The improvement of 0.0357 over IE-Critic-CoT demonstrates that reinforcement learning effectively enhances the model's alignment with human assessment and improves its generalization capability.
We also evaluate our two-stage training framework on AGIQA-3k~\cite{agiqa}, shown in Table~\ref{tab:agiqa3k}. For fair comparison, all the models are trained on the AGIQA-3k trainset. The CoT SFT data is generated by GPT-4o, using similar method. These results show the effectiveness of our proposed methods. RLVR is a good way to improve the performance and generalization.

\begin{table}[t]
\centering
\caption{Quantitative comparison of different methods on AGIQA-3k~\cite{agiqa} validation set.}
\label{tab:agiqa3k}
\resizebox{\columnwidth}{!}{
\begin{tabular}{c|ccc}
\toprule
\textbf{Method} & \textbf{PLCC} $\uparrow$ & \textbf{SROCC} $\uparrow$ & \textbf{MainScore} $\uparrow$ \\
\midrule
NIMA~\cite{nima} & 0.8208 & 0.7941 & 0.8075 \\
CLIP-IQA~\cite{clipiqa} & 0.8053 & 0.8426 & 0.8240 \\
TOPIQ~\cite{topiq} & 0.8405 & 0.8090 & 0.8248 \\
IP-IQA~\cite{IP-IQA} & 0.9116 & 0.8634 & 0.8875 \\
\midrule
Q-Align~\cite{q-align} & 0.9027 & 0.8614 & 0.8821 \\
Qwen-2.5-VL-7B~\cite{qwen-2.5-vl} & 0.9139 & 0.8804 & 0.8971 \\
\cellcolor{blue!5}+ CoT SFT & \cellcolor{blue!5}0.9096 & \cellcolor{blue!5}0.8800 & \cellcolor{blue!5}0.8948 \\
\cellcolor{blue!5}+ RLVR & \cellcolor{blue!5}\textbf{0.9307} & \cellcolor{blue!5}\textbf{0.9002} & \cellcolor{blue!5}\textbf{0.9155} \\
\bottomrule
\end{tabular}
}
\end{table}

% \subsection{Qualitative Results}
% We further conducted a qualitative comparison for different score levels in IE-Bench, as illustrated in Figure~\ref{fig:demo}, where we present several image examples in IE-Bench with varying predicted sores.

% \begin{figure}[hbp]
% \centering
% \includegraphics[width=1.0\columnwidth]{figs-tie/demo.pdf} % Reduce the figure size so that it is slightly narrower than the column. Don't use precise values for figure width.This setup will avoid overfull boxes.
% \caption{Demo of scores in IE-QA.}
% \label{fig:demo}
% \end{figure}

\subsection{Ablation Study}
To understand the key factors contributing to our model and identify the "R1 Moment" in image editing quality assessment, we conduct comprehensive ablation studies on two critical components: training recipes and reward functions. These ablations reveal the individual contributions of different design choices and validate the effectiveness of our proposed approach.

\subsubsection{Training Recipe}
We investigate different training strategies to understand how the combination of SFT data and RL training affects the final performance. Specifically, we compare the following SFT configurations:
\begin{itemize}
    \item No SFT: The model skips the SFT stage and directly undergoes RL training from the Qwen-2.5-VL-7B-Instruct model.
    \item Direct SFT: The model is trained to directly predict scores without explicit reasoning, using the base Qwen-2.5-VL-7B model with only scoring data.
    \item CoT-only SFT: The model is trained exclusively on CoT reasoning data generated by GPT-4o, where it learns to generate reasoning processes followed by score predictions.
    \item CoT + Direct SFT: The model is trained on a mixture of CoT reasoning data and direct scoring data, enabling it to learn both reasoning and direct prediction capabilities. This is the final choice to get IE-Critic-CoT.
\end{itemize}
Table~\ref{tab:ablation_recipe} presents the results of our training recipes ablation study. Several key observations emerge from these results. First, comparing the baseline Direct SFT (MainScore: 0.8208) with No SFT + RLVR (MainScore: 0.7940), we find that conducting RL without proper SFT initialization leads to degraded performance, demonstrating the importance of the cold-start phase. We also find that, RL without cold-start SFT is prone to reward hacking, where the model outputs short, fixed and useless CoT and then outputs the score. Second, CoT-only SFT without RL achieves a MainScore of 0.8029, which is lower than Direct SFT, suggesting that pure CoT training alone is insufficient. However, when combined with RLVR, CoT-only achieves 0.8446, showing significant improvement and validating the synergy between CoT reasoning and RL. 

\begin{figure}[t]
\centering
\includegraphics[width=1.\columnwidth]{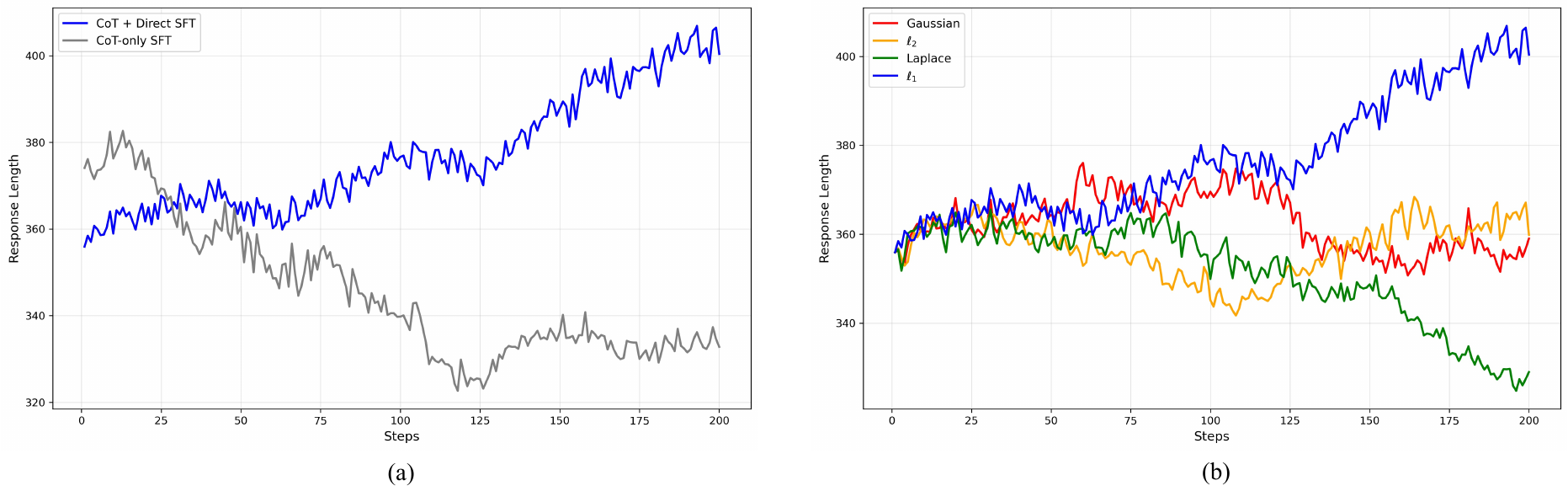} % Reduce the figure size so that it is slightly narrower than the column. Don't use precise values for figure width.This setup will avoid overfull boxes.
\caption{Response length curves during RL training. (a) CoT + Direct SFT maintains longer and more stable responses compared to CoT-only SFT. (b) The $\ell_1$ reward function achieves the best performance in encouraging detailed reasoning, while Laplacian, $\ell_2$, and Gaussian rewards show declining trends.}
\label{fig:response_length}
\label{fig:response-length-curve}
\end{figure}

Most importantly, our proposed CoT + Direct SFT mixture achieves the best results both with and without RLVR. Without RL, it reaches 0.8304, outperforming both Direct SFT and CoT-only SFT. With RLVR, it achieves the highest MainScore of 0.8661, representing a gain of 0.0215 over CoT-only + RLVR. This demonstrates that the task decomposition and data mixture strategy is crucial for enabling the model to generate stronger positive trajectories during RL, thereby raising the learning upper bound. The combination of CoT reasoning capability and direct scoring ability allows the model to flexibly adapt its output strategy, which proves essential for finding the "R1 Moment" in this domain. Figure~\ref{fig:response_length}(a) illustrates the response length curves during RL training for CoT + Direct SFT and CoT-only SFT. The CoT + Direct mixture maintains stable and continuously increasing response length throughout training, reaching approximately 400 tokens finally. In contrast, CoT-only SFT shows a declining trend, dropping from 374 tokens to 337 tokens. This observation provides important evidence that the task decomposition strategy not only improves quantitative metrics but also encourages the model to generate more comprehensive reasoning processes, which is critical for explainable quality assessment.

\begin{table}[h]
\centering
\caption{Ablation study on the training recipes.}
\label{tab:ablation_recipe}
\resizebox{\columnwidth}{!}{
\begin{tabular}{cc|ccc}
\toprule
\textbf{SFT} & \textbf{RLVR} & \textbf{PLCC} $\uparrow$ & \textbf{SROCC} $\uparrow$ & \textbf{MainScore} $\uparrow$ \\
\midrule
Direct & No & 0.8225 & 0.8191 & 0.8208 \\
No & Yes & 0.7973 & 0.7906 & 0.7940 \\
CoT-only & No & 0.8100 & 0.7958 & 0.8029 \\
CoT-only & Yes & 0.8475 & 0.8417 & 0.8446 \\
\cellcolor{blue!5}CoT + Direct & \cellcolor{blue!5}No & \cellcolor{blue!5}0.8384 & \cellcolor{blue!5}0.8224 & \cellcolor{blue!5}0.8304 \\
\cellcolor{blue!5}CoT + Direct & \cellcolor{blue!5}Yes & \cellcolor{blue!5}\textbf{0.8693} & \cellcolor{blue!5}\textbf{0.8629} & \cellcolor{blue!5}\textbf{0.8661} \\
\bottomrule
\end{tabular}
}
\end{table}

\subsubsection{Reward Function}
The choice of reward function significantly impacts the RL training process and the final model performance. We evaluate four different shaped reward functions as described in Section 4.2.2: $\ell_1$ (Linear), $\ell_2$ (Quadratic), Laplacian, and Gaussian. All functions share the same hyperparameters ($r_{\text{min}} = 0.05$ and $d_0 = 1$) to ensure fair comparison.

Table~\ref{tab:ablation_reward} presents the results of our reward function ablation study. The $\ell_1$ reward function achieves the best performance across all metrics, with a MainScore of 0.8661, outperforming $\ell_2$ by 0.005, Gaussian by 0.0055, and Laplacian by 0.0088. This superior performance can be attributed to the moderate and stable characteristics of the linear decay function. The $\ell_1$ reward provides a balance that neither over-penalizes small errors nor under-penalizes large errors, enabling the model to learn effectively across different error ranges.

Furthermore, our analysis reveals that the $\ell_1$ reward function encourages the model to generate longer and more detailed reasoning processes while maintaining high prediction accuracy. As shown in Figure~\ref{fig:response_length}(b), the $\ell_1$ reward maintains stable and continuously increasing response length throughout training, while other reward functions show significant degradation. Specifically, Laplacian experiences the most severe decline, dropping from 370 to below 330 tokens. The $\ell_2$ and Gaussian functions also exhibit declining trends. This finding validates our choice of $\ell_1$ as the reward function for IE-Critic-R1 and demonstrates the importance of careful reward design in achieving the "R1 Moment" for image editing quality assessment.

\begin{table}[h]
\centering
\caption{Ablation study on the reward functions.}
\label{tab:ablation_reward}
\resizebox{\columnwidth}{!}{
\begin{tabular}{c|ccc}
\toprule
\textbf{Reward Function} & \textbf{PLCC} $\uparrow$ & \textbf{SROCC} $\uparrow$ & \textbf{MainScore} $\uparrow$ \\
\midrule
\cellcolor{blue!5}$\ell_1$ (Linear) & \cellcolor{blue!5}\textbf{0.8693} & \cellcolor{blue!5}\textbf{0.8629} & \cellcolor{blue!5}\textbf{0.8661} \\
$\ell_2$ (Quadratic) & 0.8636 & 0.8586 & 0.8611 \\
Laplacian & 0.8593 & 0.8552 & 0.8573 \\
Gaussian &  0.8625 & 0.8586 & 0.8606 \\
\bottomrule
\end{tabular}
}
\end{table}

\section{Conclusion}
In this research, we introduce IE-Bench, a comprehensive benchmark dataset containing diverse source-prompt-target triplets with multi-dimensional human annotations across text alignment, fidelity, quality, and overall scores. Building upon IE-Bench, we propose a two-stage training framework that combines cold-start supervised fine-tuning with reinforcement learning, gaining IE-Critic-CoT and IE-Critic-R1. Through the decomposition of image editing quality assessment task, reasoning-then-scoring, we find a mix-up data recipe to get a strong foundation. Additionally, via the careful design of the reward function, we find the key elements leading to the "R1 moment". Benefiting from these aspects, IE-Critic-R1 can obtain a comprehensive, interpretable image editing quality assessment and achieve very good performance on the IE-Bench and AGIQA-3k benchmark. We hope that IE-Critic-R1 is helpful to both image editing methods and their evaluation and assessment.

{
    \small
    \bibliographystyle{ieeenat_fullname}
    \bibliography{main}
}

\appendix
\clearpage
\setcounter{page}{1}
% \maketitlesupplementary

\section{Statistics on IE-Bench}
In this section, we will detail the data composition and statistics of IE-Bench. 
We will first introduce its source data composition and show the corresponding cases for demonstration. 
Then we will further introduce the different types of editing methods selected. 
Finally, we will show the IE-Bench and related cases after subjective experiment scoring.

\subsection{Collection of source data in IE-Bench}

\begin{figure*}[t]
\centering
\includegraphics[width=2.0\columnwidth]{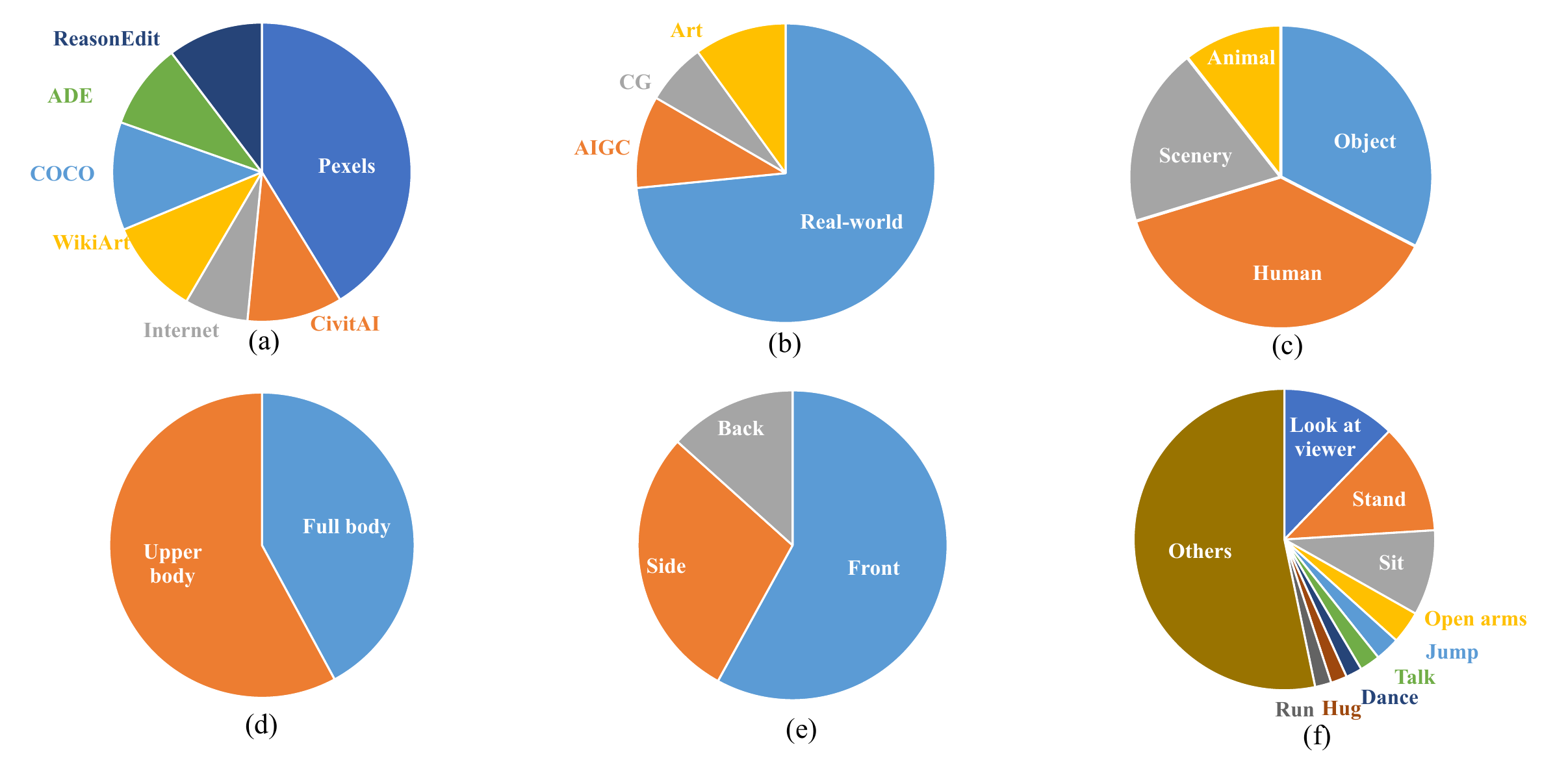} % Reduce the figure size so that it is slightly narrower than the column. Don't use precise values for figure width.This setup will avoid overfull boxes.
\caption{Collection of source images. (a) Sources of images. (b) Categories of images. (c) Content of images. (d)~(f) denotes fine-grained classification of human images, including classification of camera, pose, and actions, respectively.} %【视频数据的来源(哪些数据集)的饼状图、真实/图形学/AIGC的饼状图、Motion类别构成图、以及(人、动物、风景、物体等)内容类别构成饼状图】. }
\label{fig:src_image_collection}
\end{figure*}

The proportion of data in each part of IE-Bench can be referred to Figure~\ref{fig:src_image_collection}. When collecting data, we first confirmed that the data sources come from CG (computer-generated), AIGC (AI-generated content) images, artworks, real-world photos, etc., and identified relevant datasets. After confirming each dataset, we go through each sample, classify them according to different criteria, and apply appropriate labels. For example, images can be classified by content such as Nature, Object, Animal, Human, etc. The "Human" category is further subdivided based on the camera angle, such as half-body, full-body, and portrait, or by perspective (e.g., back view, side view, frontal view). Additionally, it can be categorized by factors like race, gender, or action type. We ensure a relatively balanced distribution through participant selection. We start by reviewing smaller datasets, prioritizing those with fewer samples, and then move on to larger datasets, selecting more diverse categories to supplement the collection. Finally, we supplement the dataset by gathering suitable content from the internet. This step is crucial. Although current datasets contain a rich variety of categories, most of the content still falls within conventional major categories. For example, landscape datasets typically include scenes like meadows or snowy mountains, but less common scenes like auroras, lava flows, and lightning are underrepresented. Additionally, despite the richness of current action classification datasets, there are still relatively few actions with large differences in pose or style. Therefore, manual screening and supplementation are necessary.

\begin{figure*}[htp]
\centering
\includegraphics[width=1.7\columnwidth]{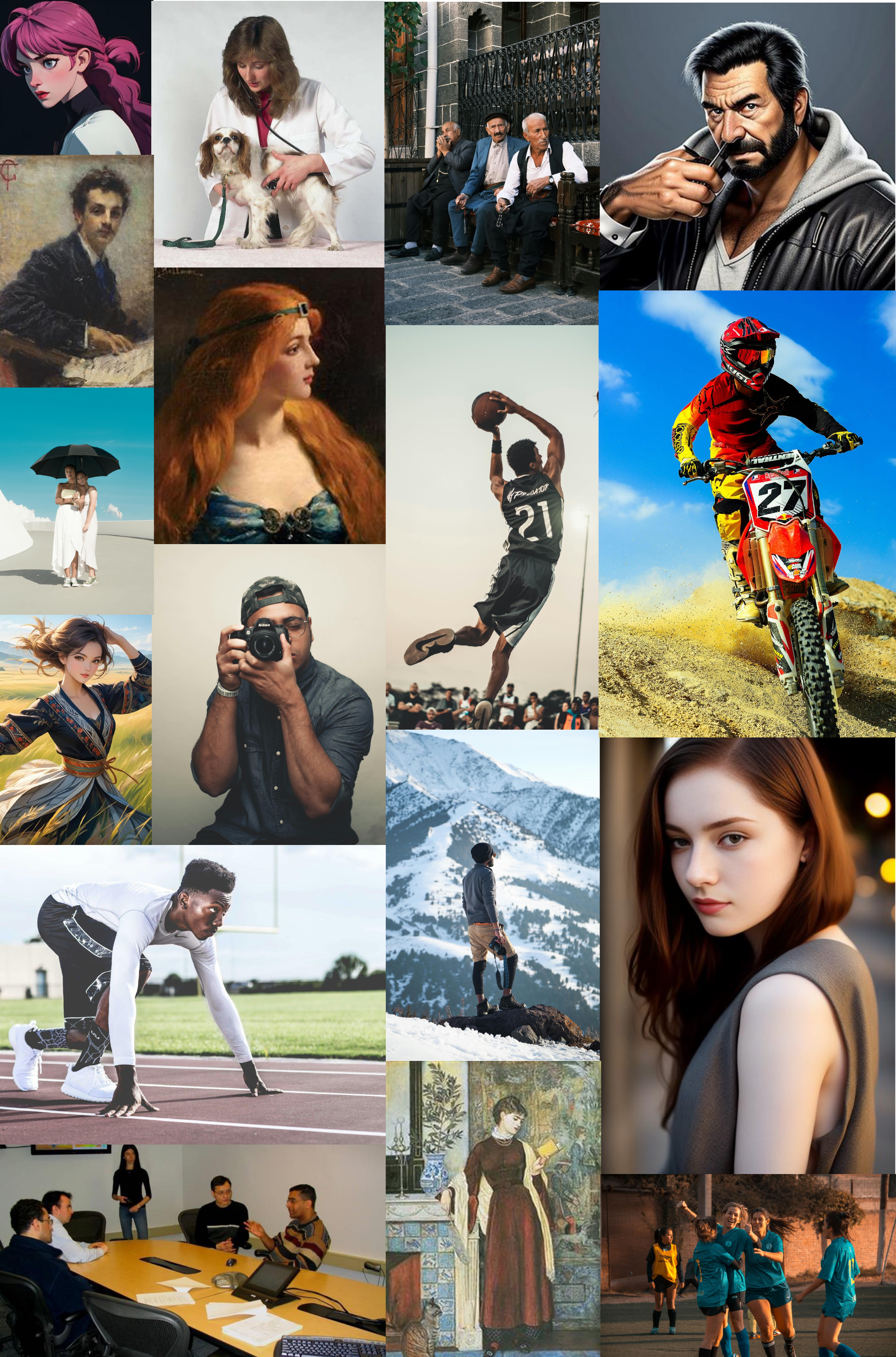} % Reduce the figure size so that it is slightly narrower than the column. Don't use precise values for figure width.This setup will avoid overfull boxes.
\caption{Examples in the collected human images.}
\label{fig:demo}
\end{figure*}

Ultimately, our dataset contains over 100 different action categories for human subjects, including actions such as looking at the viewer, sitting, crossing arms, smiling, hand on hip, hand in pocket, holding a bag, raising one hand, looking forward, turning around, hands on head, twisting hands, looking at pictures, arms folded, talking, combing hair, sitting in a bathtub, playing karate, playing golf, holding a club, looking at a golf ball, dancing, thinking, crossing arms, raising hands, opening arms, blocking sunlight with hands, looking upward, looking at each other, playing with fireworks, holding a pigeon, opening mouth, pointing with a finger, attacking, raising a fist, squatting, holding a book, playing guitar, putting hands and feet on the floor, jumping, opening legs, bowing, hugging, holding arms out, putting hands on face, holding an umbrella, skateboarding, running, holding a ball, dragging, leaning, bending the body, shooting, taking pictures, closing eyes, riding, playing tennis, bending knees, throwing a ball, boating, doing yoga, drinking, climbing, squatting, putting hands on the floor, giving a high five, pushing up, hurdling, typing, playing basketball, playing soccer, kicking the ball, etc., as shown in the Figure~\ref{fig:demo}.

Sampling is also a good method for constructing datasets. One reason we did not sample from large-scale datasets is that the video resolution in large datasets, is often lower. Such samples contain limited detail and differ significantly from the viewing habits of the human eye, which can affect subjective experiments. In addition, to achieve an appropriate proportion across multiple classification dimensions and to collect as many action categories as possible, multiple rounds of sampling and manual screening are required. Therefore, this work ultimately chose to start with small datasets and manually screen samples, ensuring both efficiency and quality of the dataset.

\begin{figure*}[t]
\centering
\begin{tcolorbox}[
    enhanced,
    colback=white,
    colframe=black,
    coltitle=white,
    colbacktitle=blue!50!cyan,
    title=Prompt for Rollout Stage of IE-Critic-R1,
    fonttitle=\bfseries\large,
    boxrule=0.8pt,
    arc=10pt,
    attach boxed title to top center={yshift=-2mm},
    boxed title style={
        enhanced,
        arc=10pt,
        top=1mm,
        bottom=1mm,
    }
]

\textbf{\textless image\textgreater\textless image\textgreater} \\

\textbf{Edit Instruction:} \{edit\_instruction\}

\medskip

Given the source image and the edited image, along with the editing instruction, please evaluate the overall quality of the edited image.

Consider the following aspects:

1. \textbf{Text alignment:} How well does the edit follow the instruction?  \\
2. \textbf{Fidelity:} How well does the edited image preserve the unedited parts of the source image?  \\
3. \textbf{Quality:} What is the perceptual quality of the edited image?

\medskip

The overall rating should be a float between 1 and 5, with 1 representing very poor quality and 5 representing excellent quality.

Return your final answer as a number rounded to two decimal places.

\medskip

A conversation between User and Assistant. The user asks a question, and the Assistant solves it. The assistant first thinks about the reasoning process in the mind and then provides the user with the answer. The reasoning process and answer are enclosed within \textless think\textgreater \textless /think\textgreater \ and \textless answer\textgreater \textless /answer\textgreater \ tags, respectively, i.e., \textless think\textgreater reasoning process here \textless /think\textgreater\textless answer\textgreater answer here \textless /answer\textgreater

\end{tcolorbox}
\caption{IE-Critic-R1 rollout prompt.}
\label{fig:rollout-prompt}

\end{figure*}

\begin{figure*}[t]
\centering
\begin{tcolorbox}[
    enhanced,
    colback=white,
    colframe=black,
    coltitle=white,
    colbacktitle=blue!50!cyan,
    title=Prompt for GPT-4o to Generate CoT,
    fonttitle=\bfseries\large,
    boxrule=0.8pt,
    arc=10pt,
    attach boxed title to top center={yshift=-2mm},
    boxed title style={
        enhanced,
        arc=10pt,
        top=1mm,
        bottom=1mm,
    }
]

\textbf{\textless image\textgreater\textless image\textgreater}

\medskip

\textbf{Edit Instruction:} \texttt{\{edit\_instruction\}}

\medskip

You will receive two images (the source image and the edited image) and the edit instruction, along with multi-dimensional scores from human annotators.

Your task is to expand the human comment comprehensively while retaining its strengths and weaknesses, making it more professional and logically rigorous. Focus only on expanding the comment and do not give or repeat any score according to the given information. Ensure the expanded comment is strictly based on the provided human comment and avoids any speculation or uncertain content.

\medskip

\textbf{[Introduction for Human Annotated Scores]}

There are four scores:
\begin{itemize}
  \item \textbf{Text alignment:} How well does the edit follow the instruction?
  \item \textbf{Fidelity:} How well does the edited image preserve the unedited parts of the source image?
  \item \textbf{Quality:} What is the perceptual quality of the edited image?
  \item \textbf{Overall:} The comprehensive score of this image editing result.
\end{itemize}

Every rating is a float between 1 and 5, with 1 representing very poor quality and 5 representing excellent quality.

\medskip

\textbf{[Multi-Dimensional Human Annotated Scores]}

\textbf{Text Alignment:} \texttt{\{normalized\_text\_score\}} \\
\textbf{Fidelity:} \texttt{\{normalized\_fidelity\_score\}} \\
\textbf{Quality:} \texttt{\{normalized\_quality\_score\}} \\
\textbf{Overall:} \texttt{\{normalized\_overall\_score\}}

\medskip

\textbf{[Expanded Comment]}

Please give me the expanded comment directly, without analysis of my user message.

\end{tcolorbox}
\caption{Prompt for GPT-4o.}
\label{fig:4o-prompt}
\end{figure*}

\subsection{Collection of methods in IE-Bench}
The editing methods we used are listed in Table~\ref{tab:editing_methods}. We employed a total of 8 models, each from different eras, architectures, optimization strategies, and editing approaches. 
These include methods ranging from early ones like Instruct-Pix2Pix to recent ones such as BAGEL, from UNet-based Stable Diffusion to DiT-based and LLM-based methods, and from approaches based on DDPM to those based on Flow Matching. Additionally, they span from inversion-based techniques to attention-manipulated methods. We further categorized these methods based on their training data. Instruction-based methods use prompts of the instruction type during training, typically requiring no full description of the generated image's elements—only highlighting the differences from the source image is necessary. In contrast, Description-based methods rely on prompts that provide detailed descriptions of the desired output.

\begin{table*}[htb]
\centering
\scalebox{1.0}{
\begin{tabular}{llccc}
\toprule
Model & Time & 0-shot & Type & Backbone. \\ \midrule
Instruct-Pix2Pix~\cite{instructpix2pix}      &  CVPR'23   &    \ding{55}   &   Instruction-based  &    SD v1-4       \\ \midrule
Prompt-to-Prompt~\cite{pnp}      &  CVPR'23    &    \ding{51}       &     Description-based       &   SD v1-5         \\ \midrule
MagicBrush~\cite{magicbrush}      &  NeurIPS'23    &    \ding{55}       &     Instruction-based       &   SD v1-5    \\ \midrule
MasaCtrl~\cite{masactrl}      & ICCV'23     &   \ding{51}          &    Description-based             &  SD v1-4          \\ \midrule
InfEdit~\cite{infedit}      & CVPR'24     &  \ding{51}           &  Description-based      &    SD v2-1       \\ \midrule
H-Edit~\cite{h-edit} & CVPR' 25 & \ding{51} & Description-based & SD v1-4 \\ \midrule
FireFlow~\cite{fireflow} & ICML' 25 & \ding{51} & Description-based & FLUX-1-dev \\ \midrule
BAGEL~\cite{bagel} & Tech Report' 25 & \ding{51} & Unified & Qwen2.5-LLM \\
\bottomrule
\end{tabular}
}
\caption{Collection of the editing models.}
\label{tab:editing_methods}
\end{table*}

\subsection{Example cases in IE-Bench}

\begin{figure*}[hbp]
\centering
\includegraphics[width=1.6\columnwidth]{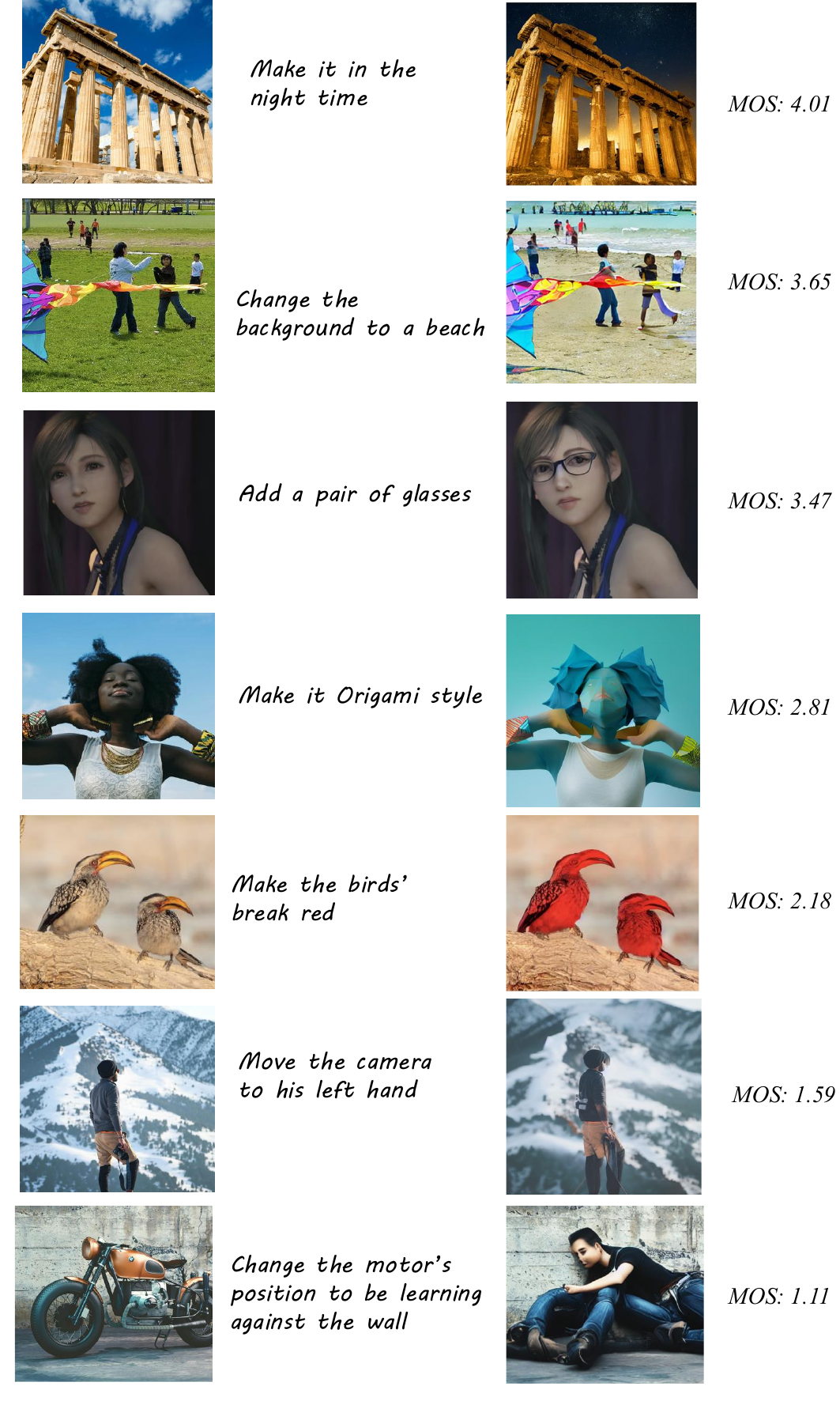} % Reduce the figure size so that it is slightly narrower than the column. Don't use precise values for figure width.This setup will avoid overfull boxes.
\caption{Example of scores for different cases in IE-Bench.}
\label{fig:demo-score}
\end{figure*}

In this section, we present the case scores collected by IE-Bench after editing and subjective experiments, as illustrated in Figure~\ref{fig:demo-score}, where several image examples from the IE-Bench with varying predicted scores are shown. These images demonstrate how the scores are collected for different images in the current set of IE-Bench.

\section{More Training Details}
In this section, we show more details about IE-Critic-CoT and IE-Critic-R1 training processes.

\subsection{IE-Critic-CoT} 
We use Qwen-2.5-VL-7B-Instruct to intialize our model and - supervised fine-tuning and choose the well-known LLaMA-Factory~\cite{llama-factory} repository as our training framework. We prompt GPT-4o to generate reasoning process CoTs, which include the text alignment, fidelity, quality scores and a summary before obtaining the overall score. For better performance and better characteristic for the subsequent RLVR, the direct scoring (without any other text) data is mixed up with the CoT data. The hyper-parameters are shown in Table~\ref{tab:sft-hpara}.

\begin{table}[h]
\centering
\caption{Training hyperparameters of IE-Critic-CoT}
\label{tab:sft-hpara}
\resizebox{\columnwidth}{!}{
\begin{tabular}{c|c}
\toprule
\textbf{Setting} & \textbf{Value}\\
\midrule
Training Epochs & 2 \\
Global Batch-size & 32 \\
Optimizer & AdamW \\
Warm-up Ratio & 0.1 \\
LR Scheduler & Cosine Decay \\
Learning Rate & 1e-5 \\
Minium Learning Rate & 1e-7 \\
Weight Decay & 0 \\
Max Gradient Norm & 1.0 \\
Image Pixels & Min: 262144 \ Max: 1048576 \\
Precision & Bfloat16 \\
DeepSpeed & ZERO-3 \\

\bottomrule
\end{tabular}
}
\end{table}

\subsection{IE-Critic-R1}
Based on IE-Critic-CoT, we adopt the Easy-R1~\cite{easyr1} framework, based on VeRL~\cite{verl}. The model is trained for five episodes using IE-Bench training data. In order to better balance exploration and exploitation, the rollout batch size is equal to train batch size, completely on-policy behavior. For more exploration and better performance, the coefficient of KL loss is set to 0. The hyper-parameters are shown in Table~\ref{tab:rlvr-hpara}.

\begin{table}[h]
\centering
\caption{Training hyperparameters of IE-Critic-R1}
\label{tab:rlvr-hpara}
\resizebox{\columnwidth}{!}{
\begin{tabular}{c|c}
\toprule
\textbf{Setting} & \textbf{Value}\\
\midrule
Training Episodes & 5 \\
Rollout Batch-size & 128 \\
Global Batch-size & 128 \\
Optimizer & AdamW \\
LR Scheduler & Constant \\
Learning Rate & 1e-6 \\
Weight decay & 0.01 \\
Max Gradient Norm & 1.0 \\
KL Coefficient & 0 \\
Image Pixels & Min: 262144 \ Max: 1048576 \\
Rollout-N & 8 \\
Temperature & 1.0 \\
Train Top-P & 1.0 \\
Validation Top-P & 0.9 \\
Format Weight & 1.0 \\

\bottomrule
\end{tabular}
}
\end{table}

\subsection{Reward Hacking w/o ColdStart}
Reward hacking refers to situations in which the model discovers and exploits unintended weaknesses or loopholes in the reward function, thereby achieving high measured reward without genuinely performing the task as intended by the system designer. As for the model without CoT SFT cold-start, we find a phenomenon where the accuracy reward grows but the reasoning CoTs are fixed and rigid. The policy model try to output the overall score directly, without accurate reasoning processes. For instance, on AGIQA-3k~\cite{agiqa} dataset, the model may output "Assessing the quality of the image involves evaluating details such as detail, realism, and overall composition" then give the finale answer. Additionally, this phenomenon occurs less frequently on IE-Bench, because image editing quality assessment needs more reasoning thoughts. Reward hacking has a great impact on the upper bound of model performance. So the cold-start stage is of great necessity and importance.

\section{Prompts}
In this section, we will provide IE-Critic-R1 rollout prompt and the prompt for GPT-4o to generate reasoing CoT. You can refer to Figure~\ref{fig:rollout-prompt} and Figure~\ref{fig:4o-prompt} for details.

% \section{Rationale}
% \label{sec:rationale}
% % 
% Having the supplementary compiled together with the main paper means that:
% % 
% \begin{itemize}
% \item The supplementary can back-reference sections of the main paper, for example, we can refer to \cref{sec:intro};
% \item The main paper can forward reference sub-sections within the supplementary explicitly (e.g. referring to a particular experiment); 
% \item When submitted to arXiv, the supplementary will already included at the end of the paper.
% \end{itemize}
% % 
% To split the supplementary pages from the main paper, you can use \href{https://support.apple.com/en-ca/guide/preview/prvw11793/mac#:~:text=Delete%20a%20page%20from%20a,or%20choose%20Edit%20%3E%20Delete).}{Preview (on macOS)}, \href{https://www.adobe.com/acrobat/how-to/delete-pages-from-pdf.html#:~:text=Choose%20%E2%80%9CTools%E2%80%9D%20%3E%20%E2%80%9COrganize,or%20pages%20from%20the%20file.}{Adobe Acrobat} (on all OSs), as well as \href{https://superuser.com/questions/517986/is-it-possible-to-delete-some-pages-of-a-pdf-document}{command line tools}.


\end{document}